\documentclass[letterpaper, 10 pt, conference]{ieeeconf}  

\IEEEoverridecommandlockouts                              

\usepackage{graphicx} 
\usepackage{amsmath} 
\usepackage{amssymb}  
\graphicspath{{media/}}

\title{\LARGE \bf
Feature-Based Transfer Learning for Robotic Push Manipulation
}

\author{Jochen St\"uber, Marek Kopicki, Claudio Zito$^{1}$
\thanks{$^{1}$Jochen St\"uber , Marek Kopicki, and Claudio Zito are with the School of Computer Science, University of Birmingham, B15 2TT, Birmingham, United Kingdom {\tt\small \{jxs1075,msk,c.zito\}@cs.bham.ac.uk}}%
}

\begin{document}

\maketitle
\thispagestyle{empty}
\pagestyle{empty}

\begin{abstract}

This paper presents a data-efficient approach to learning transferable forward models for robotic push manipulation. Our approach extends our previous work on contact-based predictors by leveraging information on the pushed object's local surface features. We test the hypothesis that, by conditioning predictions on local surface features, we can achieve  generalisation across objects of different shapes. In doing so, we do not require a CAD model of the object but rather rely on a point cloud object model (PCOM). Our approach involves learning motion models that are specific to contact models. Contact models encode the contacts seen during training time and allow generating similar contacts at prediction time. Predicting on familiar ground reduces the motion models' sample complexity while using local contact information for prediction increases their transferability. In extensive experiments in simulation, our approach is capable of transfer learning for various test objects,  outperforming a baseline predictor. We support those results with a proof of concept on a real robot.

\end{abstract}
\section{INTRODUCTION}

Pushing is an essential skill for both humans and robots. While prehensile manipulation is versatile and elegant, neither is it always possible nor suitable. Objects may be too large or heavy to grasp or the robot may not be equipped with grippers. Mobile robots in particular are posed to encounter an unpredictable variety of real-world manipulation tasks without the ability to grasp, i.e. adding a gripper to NASA's K-Rex rover means extra payload \cite{king2016rover}. Rocks on the moon may nevertheless obstruct its path. In a more mundane context, precise pushing allows industrial and personal robots to operate efficiently in cluttered environments by removing obstacles and reducing pre-grasp uncertainty \cite{dogar2012nonprehensile}. In both described contexts, robots may encounter objects of various sizes, shapes, and materials. Hence, there is a strong case for learning transferable forward models of robotic push manipulation.

However, generalisation across objects is challenging and still an open problem. Analytical approaches to pushing rely on knowledge of intrinsic physical parameters which may not be available, or expensive to obtain. Data-driven pushing models tend to either generalise poorly, require impractically large amounts of data, or both. 

We present a contact-based approach to push prediction that enables generalisation across objects of different shapes. Our aim is not to improve the quality of predictions for a specific object, but to efficiently learn a generative model that will enable the robot to make reasonable predictions on novel objects. We have also based our approach on features that can be extracted from a point cloud obtained on the fly by the robot, so to remove the need of modelling various types of objects. We do so by approximating the global motion of the object by learning a set of \textit{local experts}. Each expert is a specialised predictor learned from demonstration. 

In particular, we learn two types of experts: 1) static contact models and 2) motion models. The former models encode the contact between the pusher and the object, and between the object and the environment. Meanwhile, the latter models make predictions on how the object will behave under a specific push. 
In other words, instead of learning motions for a specific object, we break down the problem to learn how few local points of interest (i.e. robot-object and object-environment contacts) will change under a push. We then use those predictions to estimate the global motion of the object.

\begin{figure}
  \centering
  \includegraphics[clip, trim=0cm 0cm 0cm 1.8cm, width=0.99\columnwidth]{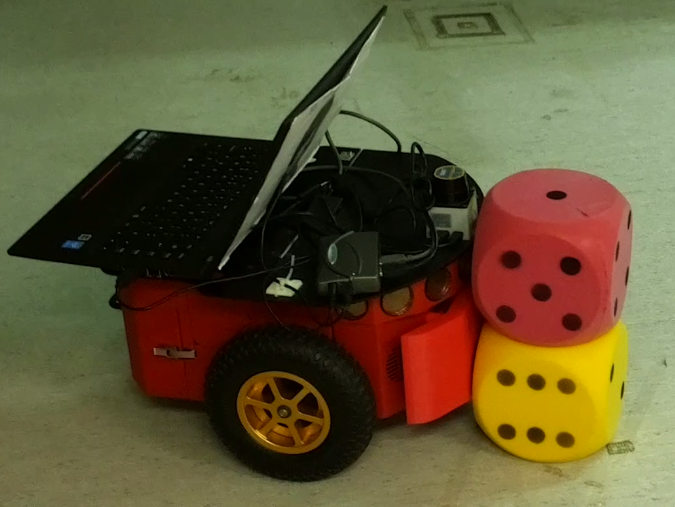}
  \caption{Our test scenario. A Pioneer 3DX equipped with a 3D printed bumper pushing a novel object. The robot is capable of predicting the effects of the push even if the dice are not present in the training data.}
  \label{fig:dice}
\end{figure}

Figure~\ref{fig:dice} shows our test scenario in which a Pioneer 3DX pushes an  object not previously encountered, composed of two piled dice. All our models have been trained in a simulated environment, also the one tested in the real scenario. To cope with the discrepancy between the simulated and the real world, we learned the motion models by rollouts with sampled friction coefficients (see Sec. \ref{sec:motion_model}).

When encountering a novel object, we use the contact model to generate robot-object and object-environment contacts similar to those seen during training, and apply the contact-specific motion models to predict how the object will behave under a push. As the generated contacts are similar to those seen during training time, the motion models can predict on familiar ground. 

Our main contributions are thus twofold. First, we propose a novel approach to transfer learning of robotic push manipulation. Second, we test that approach in extensive experiments in a simulated environment, and in a proof of concept on a real robot.

\section{RELATED WORK}

   \begin{figure}
      \centering
      \includegraphics[clip, trim=0cm 2.5cm 0cm 3cm, width=0.99\columnwidth]{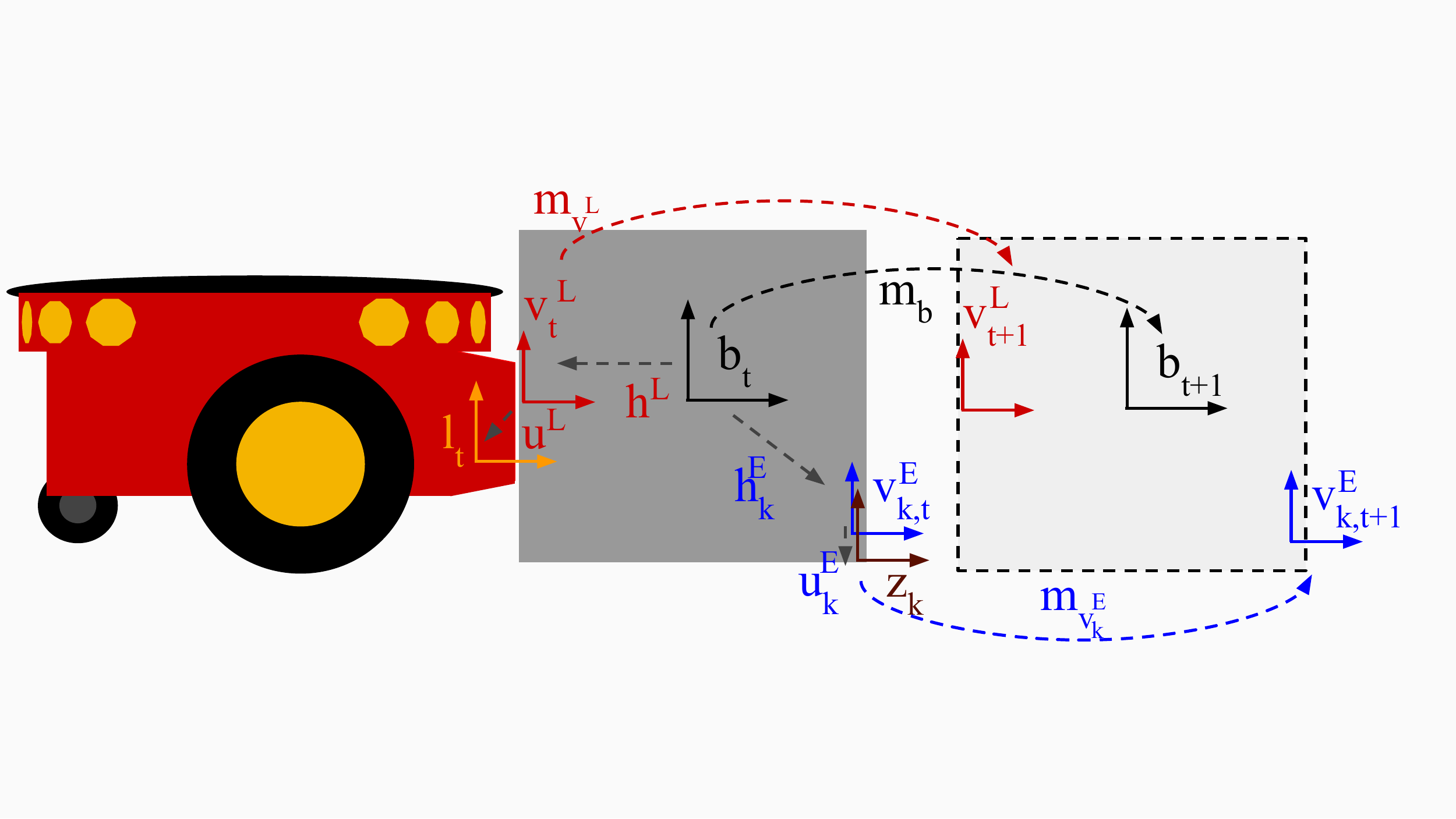}
      \caption{An object is pushed from an initial pose (dark grey) to a final pose (light grey, dotted line). We aim to predict the \textit{global} motion $m_b$ by learning \textit{local} predictors for the motions $m_{v^L}$ and $m_{v_k^E}$, for $k=1,\dots,N_E$. We represent motion, $m_x$ as a rigid body transformation that moves the frame $x_t$ into frame $x_{t+1}$. The rigid body transformations $h^L$ and $h^E_k$ describe the estimated contacts on the object's surface w.r.t. the estimated global frame of the object, $b$. This relation does not change over time, thus once we estimate the local motions $m_{v^L}$ and $m_{v_k^E}$ we use the relations $h^L$ and $h^E_k$ to estimate $b_{t+1}$. The transformations $u^L$ and $u^E_k$ denote the relative pose of the bodies in contact, respectively the robot link $l_t$ and the environment $z_k$, w.r.t. the \textit{feature} contact frames $v^L$ and $v^E_k$. This allows us to learn independent contact-specific motion models.}
      \label{fig:frames}
   \end{figure}

One approach to predicting the result of a push involves building models informed by classical mechanics \cite{mason1982manipulation, peshkin1988motionslide, lynch1992tactilefeedback}. In order to be precise, such models require explicit representation of intrinsic physical parameters such as friction. Most of those approaches focus on planar sliding of simple two-dimensional (2D) objects. Recently, there has been a promising stream of work, however, that aims to augment analytical models with data-driven approaches, e.g. for modelling the stochastic nature of friction \cite{zhou2017contactmodel}. 

In view of avoiding the tuning of physical parameters, a second approach is to learn forward models from data, thereby implicitly encoding physical parameters in the model. A wide range of features and machine learning techniques have been used for pushing prediction, e.g. recently \cite{bauza2017planarprob}. However, most approaches tend to work well only on the objects that they have been trained on, although notable exceptions exist, e.g. \cite{kopicki2017forwardmodel}. Recently, artificial neural networks have been applied to robotic pushing. Such deep learning approaches typically use RGB images as input, and may be used to construct end-to-end systems which optimise perception, prediction, and control jointly for task performance. While such work holds great promise, existing approaches require hundreds of hours of video material corresponding to thousands of actions for training to be applicable in real world scenarios \cite{agrawal2016pokelearn, finn2016unsupervisedphysics}. Generalisation across objects further increases sample complexity.

Most related to our approach is our own previous work \cite{kopicki2017forwardmodel,bib:zito_2012,kopicki2016oneshot}. In \cite{kopicki2017forwardmodel,bib:zito_2012}, we dealt with the problem of learning forward models to predict how objects behave under push operations, and in \cite{kopicki2016oneshot}, we proposed to use local features to transfer learned grasps to novel objects. In this paper, we investigate both approaches further to learn contact-specific motion models for push operations.

\section{BACKGROUND}

\subsection{Surface features}

A central aspect of our approach is the probabilistic modelling of surface features, extracted from three-dimensional (3D) object point clouds. Features are composed of a 3D position, a 3D orientation, and a 2D local surface descriptor that encodes local curvatures. Let us denote by $SO(3)$ the group of rotations in three dimensions. A feature belongs to the space $SE(3) \times \mathbb{R}^2$ , where $SE(3) = \mathbb{R}^3 \times SO(3)$ is the group of 3D poses, and surface descriptors are composed of two real numbers. We thus represent an object as a set $S$ of $N_{S}$ surface features $x_j$
\begin{equation}
S = \{x_j : x_j \in \mathbb{R}^3 \times SO(3) \times \mathbb{R}^2\}_{j\in[1,N_{S}]}.
\end{equation}
Let us denote the separation of a feature $x$ into $p \in \mathbb{R}^3$ for position, a quaternion $q \in SO(3)$ for orientation, and $r \in \mathbb{R}^2$ for the surface descriptor.  For compactness, we also denote the pose of a feature as $v$. As a result, we have $x = (v, r)$, and $v = (p, q)$.

We now describe how we compute $r$, the local curvature descriptors. The surface normal at $p$ is computed from the nearest neighbours of $p$ using a PCA-based method, e.g. \cite{kanatani2005geometry}. Surface descriptors correspond to the local principal curvatures \cite{spivak1999geometry}. The curvature at point $p$ is encoded along two directions that both lie in the plane tangential to the object’s surface, i.e. perpendicular to the surface normal at $p$. The first direction, $k_1 \in \mathbb{R}^3$ , is a direction of the highest curvature. The second direction, $k_2 \in \mathbb{R}^3$, is perpendicular to $k_1$. The curvatures along $k_1$ and $k_2$ are denoted by $r_1 \in \mathbb{R}$ and $r_2 \in \mathbb{R}$ respectively, forming a 2-dimensional feature vector $r = (r_1, r_2) \in \mathbb{R}^2$. The surface normals and principal directions allow us to define the 3D orientation $q$ that is associated to a point $p$. Next, we introduce the notation and concepts required to represent motion models.

\subsection{Rigid body motions}

Let us now consider the problem of a robotic agent with link $L$ pushing an object $B$, observed at discrete time steps $t$, $t+1$; see Fig.~\ref{fig:frames} for a visual representation of the notation. We assume that both $L$ and $B$ are 3D \textit{rigid bodies} and that the interaction between them is \textit{quasi-static}. In this context quasi-static means that the object only moves when it is being pushed by the robot with a constant low-speed velocity. Let $l_{t} \in SE(3)$ denote the pose of link $L$ and $b_{t} \in SE(3)$ the pose of $B$, measured at time $t$  in an inertial reference frame $W$ which we will refer to as the \textit{world frame}. In this paper, we aim to determine the motion of an object resulting from a pushing action $a$ from an action set $\mathcal{A}$. Let $m_b\in SE(3)$  denote the rigid body transformation from $b_{t}$ to $b_{t+1}$. Using the same notation introduced for 3D poses, we represent the separation of $m_{b}$ into $p \in \mathbb{R}^3$ for the translation, and a quaternion $q \in SO(3)$ for the rotation. Our approach is probabilistic in that we we aim to learn not a deterministic mapping $a \mapsto m_{b}$ but rather a conditional probability density function (PDF) over rigid body motions $P(m_{b}\mid a)$. 

In order to determine $P(m_{b}\mid a)$, we use a PoE. The key idea of this approach is that candidate motions are evaluated by a set of models (the experts), each giving a likelihood (opinion). For a candidate to be considered likely, all experts must agree. This is implemented by taking the product of opinions. In contrast to mixture models, each expert thus has a veto in the sense that a single zero probability results in an overall zero likelihood. 

In our context of motion prediction, we consider multiple \textit{local} predictors of the \textit{global} object motion $m_b$ (see Fig. \ref{fig:frames}). Each of those models predicts the object's local motion at point where the object is in contact with the robot or the environment. The PoE is a natural fit for this context as each contact encodes a local kinematic constraint on the object's motion. To be considered probable, a motion $m_b$ must satisfy all of those constraints simultaneously. To formalise this, let us denote a local surface feature in contact with robot link $L$ as $x^{L} = (v^L, r^L)$. We represent the contact by the pose $u^L \in SE(3)$ of $L$ relative to the local feature frame $v^L$.  In addition to the robot-object contact, we consider at set of $N_E$ contacts $u^{E}_k $ with corresponding features $x^{E}_k = (v^{E}_k , r^{E}_k )$. Let us now again consider the robot applying a pushing action $a$ to the object. In addition to $m_{b}$, we can now additionally observe the motion of each local feature frame, namely $m_{v^{L}}$ for the robot-object contact, and  $m_{v^{E}_k}$ for the object-environment contacts where $k=1,\dotsc,N_E$ .

We note that each local contact feature frame $v^L$ and $v^E_k$  resides at a relative pose $h$ to $b$ which, for a particular $v$ is given by $h = v^{-1}\circ b$, where $\circ$ denotes the pose composition operator, and $v^{-1}$ is the inverse of $v$, with $v^{-1} = (-q^{-1} p, q^{-1})$. As we assume that we are dealing with rigid bodies, $h_{t}=h_{t+1} \ \forall \ t$. For any given $v$, $m_v$ and $h$, we can thus compute the corresponding $m_b$. In other words, given the initial pose of a local contact frame, its relative pose to the object pose, and its local motion, we can compute the corresponding object motion. Coming back to our previous set-up, we learn a model for the motion of the robot-object contact frame $v^L$ given by $P(m_{v^L} \mid a)$ and a motion model for object-environment contact frames, given by $P(m_{v^{E}_k} \mid a)$ where $k=1,\dotsc,N_E$ . Given the relative poses $h^L$ and $h^E_k$, those motion models also define probability densities over $m_b$. Crucially, this formulation is transferable in that the learned motion is given in the local contact frame. When predicting the motion for a novel object, the corresponding relative poses $h^L$ and $h^E_k$ are used to generate $P(m_b \mid a)$. In the next section, we discuss how we estimate probability densities from data.

\subsection{Kernel density estimation}

In this paper, we represent PDFs non-parametrically with a set of features, or particles. The underlying PDF is created through \textit{kernel density estimation} \cite{silverman1986density}, by assigning a kernel function to each particle supporting the density. For contact models, we consider PDFs defined on surface features $x$, i.e. on $SE(3) \times \mathbb{R}^2$.  For that purpose, let us denote by $\mu$ a surface feature given by $\mu  = (\mu_p , \mu_q , \mu_r )$, and by $\sigma$ a triplet of real numbers $\sigma = (\sigma_p , \sigma_q , \sigma_r )$. We thus define our kernel as
\begin{equation}
\mathcal{K}(x \mid \mu, \sigma) = \mathcal{N}_ 3 (p\mid\mu_p , \sigma_p )\Theta(q\mid\mu_q , \sigma_q )\mathcal{N}_2 (r\mid\mu_r , \sigma_r ),
\end{equation}
where $\mu$ is the kernel mean point, $\sigma$ is the kernel bandwidth, $\mathcal{N}_n$ is an $n$-variate isotropic Gaussian kernel, and $\Theta$ corresponds to a pair of antipodal von Mises-Fisher distributions which form a Gaussian-like distribution on $SO(3)$ (for details see \cite{fisher1953sphere}). Given a set of $K$ surface features, the probability density in a region of space is then determined by the local density of the particles in that region, as
\begin{equation}
P(x) \simeq \sum_{j=i}^{K} w_j \mathcal{K}(x \mid x_j , \sigma),
\end{equation}
where $\sigma \in \mathbb{R}^3$ is the kernel bandwidth and $w_j \in \mathbb{R}^+$ is a weight associated to $x_j$ such that $\sum_{j}w_{j} = 1$. For motion models, we consider a kernel function defined over surface features and 3D motions, hence over $SE(3)\times SE(3) \times \mathbb{R}^2$. That kernels is defined as the product of $\mathcal{K}$ and the kernel function 
\begin{equation}
\mathcal{M}(m | \mu, \sigma)=\mathcal{N}_ 3 (p\mid\mu_p , \sigma_p )\Theta(q\mid\mu_q , \sigma_q ),
\end{equation}
where $m=(p, q)$ is the motion to be evaluated,  $\mu  = (\mu_p , \mu_q)$, and  $\sigma = (\sigma_p , \sigma_q)$. 

\section{PROPOSED APPROACH}
At training time, we learn a \textit{contact model} comprising both a robot-object contact model and an object-environment contact model. For each action in our action set, we then learn local motion models specific to that contact model. At prediction time, we can query any novel point cloud to find the closest set of contacts and interpolate a prediction for the object movement. 

\subsection{Contact model}

A contact model encodes the joint probability distribution of surface features and of the 3D pose of the robot's link in contact. 
At prediction time, we obtain a point cloud $O$ of the novel object from a single shot taken from a depth camera. Given a set of $N_O$ surface features $\{x_j\}_{j=1}^{N_O}$, with $x_j = (v_j, r_j)$ and $v_j = (p_j, q_j)$, a robot-object contact model $C^{L}$ is constructed from features from the object's surface. Surface features close to the link surface are more important than those lying far from the surface. Features are thus weighted, to make their influence on $C^{L}$ decrease with their squared distance to the link. Let us denote by $u_{j} = (p_{j}, q_{j})$ the pose of $L$ relative to the pose $v_j$ of the $j$\textsuperscript{th} surface feature. In other words, $u_{j}$ is defined as
\begin{equation}
u_{j} = v_j^{-1} \circ s,
\end{equation} 
where $s$ denotes the pose of $L$ in the world frame, $\circ$ denotes the pose composition operator, and $v_j^{-1}$ is the inverse of $v_j$. The contact model is then estimated as

\begin{align}
C^{L}&(u, r) \simeq \nonumber \\
&\frac{1}{Z} \sum_{j=1}^{N_{O}} w_{j}\mathcal{N}_3(p \mid p_{j}, \sigma_p)\Theta(q \mid q_{j}, \sigma_q)\mathcal{N}_2(r \mid r_j, \sigma_r),
\end{align}where $Z$ is a normalising constant, and $u = (p, q)$. 

Additionally, we learn an object-environment contact model  $C^{E}$ from $N_{C^{E}}$ samples obtained from $O$.  For each sampled feature $x_j$, we attach a frame $z_j$ to the closest point in the environment and represent the environment contact by the pose $u_j$ of $z_j$ relative to $v_j$. For the object-environment contact model, we opted for a simple binary weighting function
\begin{equation}
w_{j}=
\begin{cases}
   1  & \text{if } \lVert p_j - z_{j} \rVert < \delta_{E} \\
    0              & \text{otherwise,}
\end{cases}
\end{equation}

where $\delta_{E}$ is a cut-off distance. We then estimate the object-environment contact model using the same formulation given for the robot-object contact model. 

\subsection{Query density}\label{sec:query_density}
	
A query density results from the combination of a learnt contact model with a novel object point cloud $O$. The purpose of a query density is both to generate and evaluate poses of the corresponding robot's link (or the environment) on the new object. Imagine you have only learned how to push a cube from one of its side faces. When you need to push a triangular prism you will not place your finger on one of the corners, expecting it to move like the cube. However, if you place your finger on a side face you may be able to use your experience on the cube to make a prediction.
We refer the reader to \cite{kopicki2016oneshot} for further details on how to compute the query density, providing only a brief overview here. In summary, a query density $Q$ is a PDF defined as
%
%
   
\begin{equation}
Q=P(s, u, v, r),
\end{equation}where $v$ denotes a point on the object’s surface, expressed in the world frame, $r$ is the surface curvature of such a point, $u$ denotes the pose of a body relative to a local frame on the object, and $s$ is the absolute pose in the world frame of the body. In our context, the body is either the robot's link or a local environment frame.  
   
 At prediction time, we use a \textit{robot-object} query density $Q^L$ to generate poses of the robot's link on the new object, and to attach a robot-object contact frame as an expert for prediction. We achieve the former by marginalising with respect to $u$, $v$, and $r$, obtaining the distribution $Q^L(s)$ which models the pose $s$ in the world frame of the link $L$.  We approximate the robot-object query density by $K_{Q^L}$ kernels centred on the set of weighted robot link poses obtained from the pose sampling algorithm proposed by \cite{kopicki2016oneshot}. To generate a robot-object feature frame $v^L$, we choose it such that it maximises $Q^L(v)$. For optimisation we use simulated annealing \cite{kirkpatrick1983optimization}. 

In addition to the robot-object query density, we create an \textit{object-environment} query density by combining the object-environment contact model with $O$. While we are free to position the robot in the world, we consider the state of the environment as fixed for a given time $t$. We hence use the object-environment query density only to select the set of $N_E$ query object-environment contact frames $v^E_k$. This is equivalent to marginalising $Q^E$ with respect to $s$, $u$, and $r$ to obtain the distribution $Q^E(v)$ over poses in the world frame of local feature frames on the object's surface. We then sample from $Q^E(v)$ for a total of $N_E$ times. 


\subsection{Motion model}\label{sec:motion_model}

Having learned a contact model, we then proceed to learning both a robot-object motion model, and an object-environment motion model for each action $a$ in the action set. We learn those models from the data by observing the local motion of robot-object and object-environment frames at training time. For each action $a$, we simulate a set of $N_a$ rollouts which are the kernels of the corresponding motion model. Each rollout has a different friction coefficient for the contacts uniformly sampled from a distribution in the range $[0.15, 0.35]$, similar to the approach taken by \cite{zhou2017contactmodel}.  

In doing so, we learn to predict motion conditional on both the contact pose $u$ and the local curvature $r$ in the form of PDFs  $P(m_{v^L}\mid u^L, r^L, a)$ and $P(m_{v^{E}_k} \mid u^E_k, r^E_k, a)$. At prediction time, we use the relative poses $h^L$ and $h^E_k$ of the contact frames with respect to the object pose $b$ to define PDFs over object motion P($m_b \mid u^L, r^L, a)$ and  $P(m_b \mid u^E_k, r^E_k, a)$ from the local motion PDFs, as discussed before. Each kernel in the PDF of each motion model is defined over $SE(3)\times SE(3) \times \mathbb{R}^2$. In order to generate predictions, we obtain an opinion on candidate motions from each local expert defined as the conditional PDF

\begin{equation}
M(m_b \mid c, a) = \frac{\sum_{j=i}^{N_a} w_j \mathcal{M} (m_b\mid m_j , \sigma_{m} )\mathcal{K}(c \mid c_j , \sigma_c)}{\sum_{j=i}^{N_a} w_j  \mathcal{K}(c \mid c_j , \sigma_c)},
\end{equation}where $m_b=(p_{m_b}, q_{m_b})$ is the candidate object motion to evaluate, $c = (u, r)$ is the relative pose and surface curvature associated with the conditioning contact frame, $m_j$ is the $j$\textsuperscript{th} motion kernel, $\sigma_m$ is the motion kernel bandwidth parameter, and $\sigma_c=(\sigma_p, \sigma_q, \sigma_r)$ is the contact feature bandwidth parameter. Finally, we combine all local motion models in a PoE, as introduced in the background section, and select the best prediction by obtaining the argument with the maximum likelihood. For that purpose, we again use simulated annealing as a convenient optimisation procedure. 

\section{EXPERIMENTS}\label{sec:experiments}

We evaluated our approach empirically in experiments on a simulated and a real Pioneer 3DX mobile robot equipped with a bumper for pushing objects. Our experiments focus on object transfer, the main aspiration of this paper.  We learned contact and motion models in simulation for one object (i.e. the cube), and then evaluated the predictive performance of the learned models on a set of five test objects in simulation, and two test objects in a real world scenario. We also compared our approach to a baseline predictor which encodes the planar sliding constraints (Sec. \ref{sec:baseline}).

For our experiments, we varied the trained models along two dimensions, first the \textit{contact information} used for prediction, and second the \textit{number of training samples}. We considered three values in each dimension and all combinations between them, leading to a total of nine evaluated models. During both training and testing, all objects were placed in an obstacle-free planar environment. Due to the geometry of the objects considered, and the kinematics of our robot, our experiments are a study of planar sliding behaviour. Investigating performance in the context of rolling and toppling, as well as in cluttered environments, is future work. For all simulation experiments, we used the Gazebo simulator in version 7.8 with the Open Dynamics Engine (ODE). 
\subsection{Training set}

As our training object, we chose a cube with side length $0.2$ m and mass $0.5$ kg. Using a Kinect depth camera, we took a single shot of the object, and learned \textit{contact models} from the thus obtained point cloud. Specifically, we learned two robot-object contact models. Due to the symmetry of the bumper, we only distinguish between front and side link without differentiating between left and right. 

In the case of the first \textit{robot-object contact model}, the bumper's front link is in full contact with one face of the cube. In the case of the second robot-object contact model, the bumper's side link is in full contact with a face of the cube. In both cases, we used a cut-off distance of $1$ cm for the weighting function. Furthermore, we learned an \textit{object-environment contact model} from $1,000$ surface features sampled from the point cloud. Here, we used a cut-off distance of $5$ cm for the weighting function. We note that in our set-up devoid of obstacles and clutter, the object's environment contacts are exclusively made with the ground.  

In the next step, we learned a \textit{motion model} for each combination of action and robot-object contact model. To that end, we generated training pushes as follows. For each robot-object contact model,  we constructed a query density from which we generated $100$ feasible training contacts, discarding infeasible instances that placed the robot in mid-air or in collision with either the object or the environment. At each feasible contact, we then repeatedly applied a set of pushes.  Specifically, we considered an \textit{action set} of three pushes, one straight linear push and two angular pushes. In all cases, a velocity command was fed to the robot's controller for a fixed duration. We used a duration of four seconds. Linear velocity amounted to  $0.1$ m in the $x$\nobreakdash-direction of the robot's base frame for all pushes. For the straight linear push, all components of the angular velocity vector were set to the zero, while for the two angular pushes, we considered angular velocities of $10^{\circ}/s$ and $-10^{\circ}/s$ respectively, directed around the $z$\nobreakdash-axis of the robot's base frame. In the case of the side link's contact model, we excluded the angular push directed away from the contact surface from the action set due to its resulting in a loss of contact without applying a push. 

At each training contact, we executed each action five times (i.e. rollouts, Sec. \ref{sec:motion_model}), gathering a set of  $2,500$ training pushes, $1,500$ for the front link's contact model, and $1,000$ for the side link's contact model. 

\subsection{Test set}

We evaluated the learned motion models both in simulation and on a real robot. Considering the \textit{simulated environment} first, we generated a \textit{test set} of pushes covering five objects, one being the cube seen during training, and the other four being unseen objects. This design reflects our focus on object transfer. The unseen objects comprise a cuboid with side lengths $0.2$ m and $0.3$ m, an equilateral triangular prism with side length $0.2$ m, a cylinder of radius $0.1$m and height $0.2$ m, and a shape derived from the equilateral triangular prism by rounding off one of its edges. We will refer to the latter object as a rounded triangular prism. All considered objects have the same mass of $0.5$ kg.  We chose the objects such that they exhibit variance with respect to the shape, area, and curvature of their surfaces which in turn determine the nature of the contacts that can be generated.   

We generated \textit{test pushes} for the selected objects following the same process used to gather training data. For each test object, we obtained a single shot from a Kinect depth camera. For each of the two learned contact models, we constructed a query density and sampled  $50$ query poses.  For each query pose, we then applied each action from the corresponding action set four times. For each test push, we sampled the contact friction coefficients from  the same distribution used during training.  Hence, we obtained a test set of $1,000$ pushes per training object, $600$ for the front link's contact model and $400$ for the  side link's contact model.  Considering all objects, our test set thus comprised $5,000$ pushes. For \textit{model selection}, we additionally generated a separate \textit{validation set} of $1,000$ pushes. We tuned the kernel bandwidth parameters on the validation set and kept them constant across all experiments. 

Regarding, the evaluation on the \textit{real robot}, we considered a test set of two boxes. The first, with length $0.35$ m, width $0.16$ m, and height $0.15$ m, we will refer to as the \textit{large box}. Further, we will refer to the second, with length $0.22$ m, width $0.17$ m, and height $0.12$ m, as the \textit{small box}. Again, we obtained point clouds of the objects from single shots of a Kinect camera. Subsequently, we estimated each object's initial pose from the point cloud. First, we approximated the position of its centre of mass (COM) as the centroid of the point cloud. Then, we computed the object's orientation from the eigenvectors of the point cloud's covariance matrix. After applying the push, we obtained the object's final pose from a new shot of the depth camera using the same approach. As in simulation, we considered both contact models (front and side), and generated test contacts from the query density. However, we pruned the angular pushes from the action set, considering only the linear case. Following the described process, we generated $40$ test pushes for this initial proof of concept. We will extend the scope of those experiments in prioritised future work.

   \begin{figure}
      \centering
      \includegraphics[clip, trim=0cm 0cm 1.7cm 0cm, width=0.99\columnwidth]{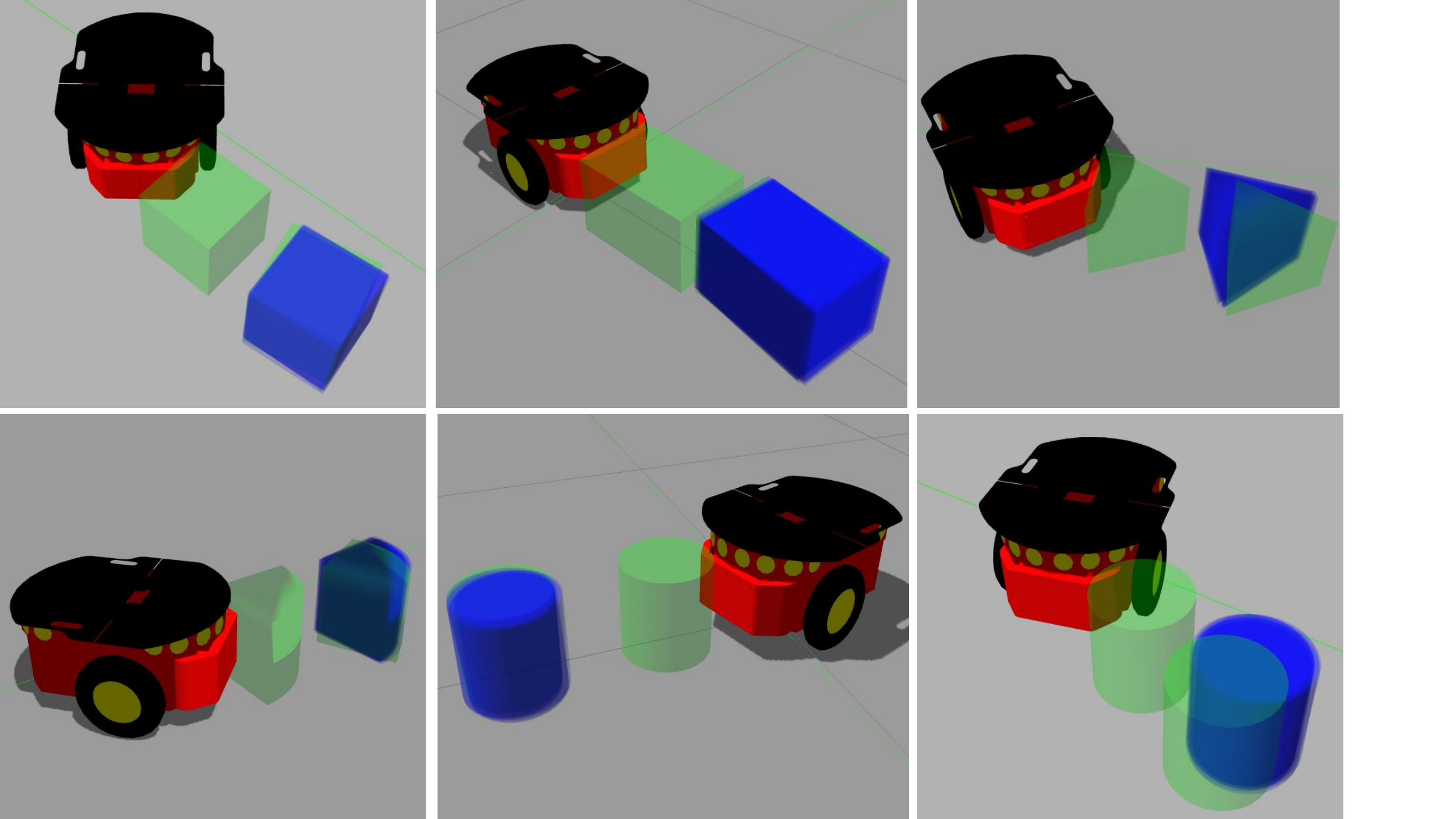}
      \caption{Pushes with predictions: initial object pose (green, in contact with robot), true final object pose (green, displaced), and predictions (blue).}
      \label{fig:pushes}
   \end{figure}

\subsection{Predictions}

We now proceed to describe how we generated predictions. For each generated test contact, we attached a set of frames. First, we attached a \textit{global object frame} at the object's estimated COM. Both in simulation and on the real robot, we estimated the location of the COM as the centroid of the point cloud. 

Subsequently, we determine the robot-object and the object-environment contact frames by using the query density as described in Sec. \ref{sec:query_density}.
In our experiments, we also aim to investigate the benefit of learning object-environment contacts to improve predictions. Hence, in our results, we denote the predictor based on only robot-object contact information as \textit{RO}, and the predictors with additional access to three and five object-environment contacts as \textit{RO+3OE}, and \textit{RO+5OE} respectively.  

Additionally, we varied the \textit{number of training pushes} available to our models. In simulation, we considered the cases of $100$, $200$, and $500$ training pushes; on the real robot, we considered $500$ pushes only. We thus trained nine different models, namely each of \textit{RO}, \textit{RO+3OE}, and \textit{RO+5OE} with the three considered training set sizes. Based on the available input information, and number of samples, each predictor generated $500$ candidate object motions for each test contact using simulated annealing. Every one of those candidates was generated as the likeliest of $100$ samples obtained from the probability density over object motions, and then optimised in $100$ iterations of the simulated annealing algorithm. Of the $500$ optimised candidates, we kept the $10$ predictions with the maximum likelihood scores. Finally, we computed error statistics for all predictions and compared them to a baseline predictor. We describe our methodology used thereby next.

   \begin{figure*}
      \centering
      \includegraphics[clip, trim =1cm 8.7cm 2cm 3cm, width=\textwidth, height=7cm]{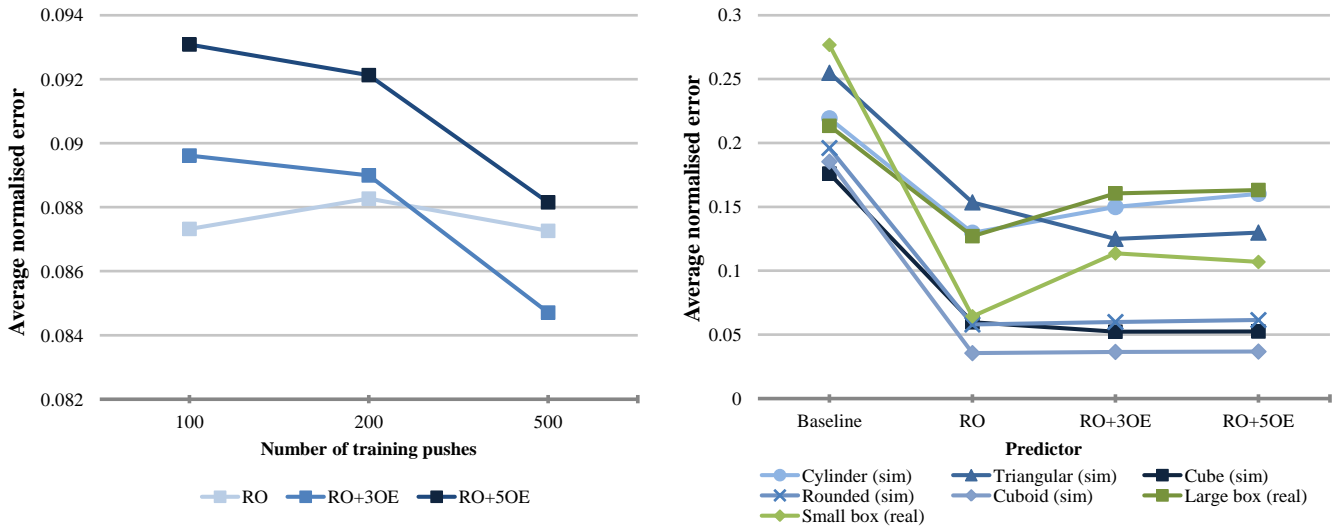}
      \caption{\textit{Left}: Performance of predictors in simulation, by size of training set. \textit{Right}: Performance of predictors trained on $500$ training pushes, by object.}
      \label{fig:naf}
   \end{figure*}

\subsection{Baseline predictor}\label{sec:baseline}

As a performance baseline for comparison, we used a \textit{baseline predictor} to all test pushes. Although simple, this predictor has several advantages over our proposed models. First, it knows the object's true COM. Foremost, though, it encodes the planar sliding constraints of our set-up. Thus, the predictor knows that objects will not topple or roll, and that they cannot penetrate the ground. In contrast, our contact-based predictors need to learn those constraints from the data. 

We implemented it such that for the $i$\textsuperscript{th} action from the action set, it applies a translation $b$ to the initial pose in world coordinates of the object's true COM, to which the predictor has access. For the linear push that translation $b$ is computed by transforming the translation given by $a_{i}\cdot l$ from the robot's base frame into world coordinates. Here, $a_{i}$ is the linear velocity vector of the velocity command corresponding to the $i$\textsuperscript{th} action, and $l$ is the action duration.  For angular pushes, we followed the same process but instead of $a_{i}$ compute the translation from the vector $\tilde{a}_{i}$ which is the vector in the $xy$\nobreakdash-plane of the robot's base frame that lies at an angle $\alpha$ to $a_{i}$, where $\alpha$ is the angular velocity around the $z$\nobreakdash-axis of the robot's base frame corresponding to the $i$\textsuperscript{th} action.   

\subsection{Performance measure}

For each prediction, we computed both a linear and an angular distance metric with regard to the final object pose. For linear distance $d_{lin}$, we opted for Euclidean distance. For angular distance $d_{ang}$, we used a quaternion distance metric defined as

\begin{equation}
d_{ang} = 1 -  (q_{test} \cdot q_{pred})^{2} ,
\end{equation}

where where $q_{test}$ and $q_{pred}$ are unit quaternions representing the final orientation in world coordinates of the object in test sample, and prediction respectively. To obtain a unified distance measure, we combine $d_{lin}$ and $d_{ang}$ by computing the normalised error $d_{norm}$ as

\begin{equation}
d_{norm} = \frac{1}{2}d_{ang} + \frac{1}{2L}d_{lin},
\end{equation}where $L$ is a normalisation constant. We set  $
L = \lVert a_{lin} \rVert \cdot l
$ where $\lVert a_{lin} \rVert$ is the Euclidean norm of the linear velocity vector $a_{lin}$ representing the linear push from our action set, and $l$ is the duration of the push applied during experiments. We chose that normalisation constant because it constitutes a critical  parameter of the problem. All else equal, the longer the push, the more challenging the prediction. In our specific set-up, $L=0.4$. 

   \begin{table*}
      \caption{Average normalised error, models trained on $500$ training pushes}
      \label{tab:naf}
      \begin{center}
      \begin{tabular}{l c c c c}
      \hline
      Object & Baseline & RO & RO+3OE & RO+5OE \\
      \hline 
      \textit{Simulation} & & & & \\
      \hline
      Cuboid & $0.185 \pm 0.126$ & $0.035 \pm 0.041$ & $0.036 \pm 0.035$ & $0.036 \pm 0.036$ \\
      Cube & $0.176 \pm 0.127$ & $0.059 \pm 0.081$ & $0.052 \pm 0.067$ & $0.052 \pm 0.067$ \\
      Rounded & $0.196 \pm 0.141$ & $0.057 \pm 0.075$ & $0.059 \pm 0.068$ & $0.061 \pm 0.071$ \\
      Triangular & $0.254 \pm 0.130$ & $0.153 \pm 0.086$ & $0.124 \pm 0.095$ & $0.129 \pm 0.097$ \\
      Cylinder & $0.219 \pm 0.111$ & $0.129 \pm 0.068$ & $0.149 \pm 0.111$ & $0.160 \pm 0.119$ \\
      \textit{Total} & $0.206 \pm 0.131$ & $0.087 \pm 0.085$ & $0.084 \pm 0.091$ & $0.088 \pm 0.096$ \\
      \hline
      \textit{Real robot} & & & & \\
      \hline
      Large box & $0.213 \pm 0.098$ & $0.127 \pm 0.027$ & $0.161 \pm 0.053$ & $0.163 \pm 0.054$ \\
      Small box & $0.276 \pm 0.077$ & $0.064 \pm 0.054$ & $0.113 \pm 0.071$ & $0.106 \pm 0.076$ \\
      \textit{Total} & $0.244 \pm 0.093 $ & $0.096 \pm 0.052$ & $0.137 \pm 0.066$ & $0.135 \pm 0.072$ \\
      \hline
      \end{tabular}
      \end{center}
   \end{table*}

\subsection{Results}

In simulation and in experiments with the real robot, all of the trained models outperformed the baseline predictor for all test objects with regard to the average normalised error. The lowest overall error value across contact information and sample sizes, with a magnitude of $0.084$, was achieved by \textit{RO+3OE} with $500$ test samples (see Fig.~\ref{fig:naf}). Breaking this down into its non-normalised components, it is equivalent to an average linear error of $6.171$ cm, and an average angular error of $0.031$. All trained models furthermore exhibited a lower standard deviation of the normalised error than the baseline predictor. Nevertheless, we see great variability in predictions with some of the standard deviations similar in magnitude to the average error itself.

Concerning the \textit{size of the training set} (Fig. \ref{fig:naf}, left), our data indicates a relationship between model complexity and sample complexity. For $100$ training samples, the simplest model \textit{RO} performs best while \textit{RO+5OE} fares worst. As the number of training pushes increases, the performance of \textit{RO} remains approximately flat. Meanwhile,  the error values for \textit{RO+3OE} and  \textit{RO+5OE} decrease.  At $500$ pushes, \textit{RO+3OE} surpasses \textit{RO} as the top performer while while \textit{RO+5OE} reaches a similar performance level as \textit{RO}.  

Considering differences between \textit{objects} (Fig. \ref{fig:naf}, right), we find that, overall, the object-environment contacts do not significantly improve  the quality of predictions, besides for the triangular prism. In some cases, discordant outputs from the local predictors diminished performance, which can happen with the PoE.  Nevertheless, the predictions are still more accurate than the baseline predictor. 

\section{DISCUSSION \& FUTURE WORK}

The experimental results obtained in simulation provide empirical support for the hypothesis that our approach enables \textit{object transfer}. For objects with sufficiently similar contact geometry, performance is comparable or even better than for the training object. Objects with very different curvatures (the cylinder), and a different area of the support surface (the triangular prism) challenge our predictors. In the first case, the query density is unable to generate sufficiently similar contacts. In the latter case, although similar contacts can be generated, the dynamics of the test object are too different from those encoded by the motion model. In the case of the rounded triangular prism, predictions are accurate as the query density is able to generate contacts on the flat faces while the support surface is large enough to result in motions similar to those of the training cube. One possible approach to increasing prediction accuracy further is to try to increase the generalisation capabilities of the learned models. A more simple, yet promising alternative is learning additional contact and motion models, and selecting the most suitable ones at prediction time. We expect that learning one model for flat surface and one for curved surfaces alone will greatly increase prediction accuracy. 

Regarding the high \textit{variability of prediction accuracy}, we find a large difference in performance between the two contact models. For the front link's contact model, the average normalised error across all models is $0.072$, compared to a value of $0.113$ for the left link's contact model. Further analysis of the data shows that for contacts with the left link, depending on the query pose and contact friction, the object came in contact with the robot's wheels, leading to highly variable results. This is a problem that can be more readily addressed through improved robot design than through modifications of the predictors.

Another reason for variable prediction performance lies in the differing quality of generated \textit{query poses}. Not all generated query poses are equally similar to the contacts seen during training time. Hence, predictions may require more or less generalisation capability depending on the query pose. Investing more effort into generating optimised query poses may thus be worth the cost depending on the achieved performance yield. Investigating this question is an interesting route for future work. 

With regard to performance gains achieved by adding further \textit{contact information}, our results are mixed. Overall best performance is achieved by \textit{RO+3OE} but \textit{RO} performs better for small training sets. Even in the case of $500$ training samples, for some objects, adding contact information indeed worsens performance. To illuminate those results, we note that combining object-environment contacts in a PoE entails the risks of obtaining small numbers. In the extreme case, this may result in numerical instability or predictions of barely diminishable likelihoods. When few kernels support the density, likelihoods are more prone to be small than in the case of abundant training data. Our findings that \textit{RO+3OE} and \textit{RO+5OE} improve with the size of the training set are consistent with those observations. Whether this explains all of the observed differences, and whether their accuracy will continue to increase at a similar rate with the number of training pushes, are questions which we aim to address in future work by utilising our full training set. 

In the light of this discussion, a particularly interesting result of our experiments is that additional contact information worsens performance for the cylinder in simulation, and for the large and small boxes in experiments with the real robot. Notably, in all cases the objects exhibit surface features which are comparably dissimilar to those seen during training time: the cylinder because of its rounded shape, the test boxes because real point clouds are noisier than the simulated ones seen during training. As motion candidates' likelihoods are evaluated conditional on surface features, this may exacerbate the aforementioned challenge of small numbers. At this point, we emphasise that this paper represents a first investigation into the proposed approach, and rigorously scrutinising the posed questions will require further experiments which we will conduct in future work. In particular, we aim to train models on real point cloud data to evaluate whether they fare better in predicting real pushes from noisy point clouds.

With regard to \textit{limitations} of our approach, we already noted that our experiments focus on planar sliding, and have primarily been conducted in simulation. Complementing this with further experiments on real robotic platforms, and with objects that are free to topple and roll, are prioritised items for future work. Alleviating the requirement of learning a separate motion model for each action would furthermore greatly increase the range of applications for our approach. Evaluating the predictive performance of alternative surface features, such as surface material, is another interesting route for further investigations. So is using information obtained from force-torque sensors for prediction. Furthermore, we are interested in learning dynamic contact models, i.e. predicting how contacts will change over time. Finally, we aim to use the presented feature-based predictors as transition models for reinforcement learning.  

\addtolength{\textheight}{-12cm}   





\bibliographystyle{abbrv}
\bibliography{bib/sources}

\end{document}